\documentclass{article}
\usepackage{ijcai24}

\usepackage{times}
\usepackage{soul}
\usepackage{url}
\usepackage[hidelinks]{hyperref}
\usepackage[utf8]{inputenc}
\usepackage[small]{caption}
\usepackage{graphicx}
\usepackage{amsmath}
\usepackage{amsthm}
\usepackage{booktabs}
\usepackage{algorithm}
\usepackage{algorithmic}
\usepackage[switch]{lineno}
\usepackage{graphicx} 
\usepackage{graphicx}
\usepackage{amsmath}
\usepackage{amssymb}
\usepackage{booktabs}

\usepackage{amsmath,amssymb,amsfonts}
\usepackage{graphicx}
\usepackage{textcomp}
\usepackage{xcolor}
\usepackage{flushend}
\usepackage{booktabs}
\usepackage{microtype}
\usepackage{graphicx}
\newcommand{\mathbbm}[1]{\text{\usefont{U}{bbm}{m}{n}#1}} 
\usepackage{booktabs} 
\usepackage{resizegather}
\usepackage{subcaption}
\usepackage{multirow}

\DeclareMathOperator*{\argmax}{arg\,max}

\def\rvx{{\mathbf{x}}}
\def\rvy{{\mathbf{y}}}

\title{Prompt-Driven Feature Diffusion for Open-World Semi-Supervised Learning}

\author{
Marzi Heidari
\and
Hanping Zhang
\and
Yuhong Guo
\affiliations
School of Computer Science, Carleton University, Ottawa, Canada
}

\begin{document}

\maketitle

\begin{abstract}
In this paper,
we present a novel approach termed Prompt-Driven Feature Diffusion (PDFD) 
within a semi-supervised learning framework
for Open World Semi-Supervised Learning (OW-SSL).
At its core, PDFD deploys an efficient feature-level diffusion model
with the guidance of class-specific prompts
to support discriminative feature representation learning and feature generation, 
tackling the challenge of the non-availability of labeled data for unseen classes in OW-SSL. 
In particular, PDFD utilizes class prototypes as prompts in the diffusion model, 
leveraging their class-discriminative 
and semantic generalization
	ability to condition and guide the diffusion process
	across all the seen and unseen classes. 
Furthermore, PDFD 
incorporates a class-conditional adversarial loss 
for diffusion model training, 
ensuring that the features generated via the diffusion process 
can be discriminatively aligned with the class-conditional features of the real data. 
Additionally, 
the class prototypes of the unseen classes are computed using only 
unlabeled instances with confident predictions
within a semi-supervised learning framework. 
We conduct extensive experiments to evaluate the proposed PDFD.  
The empirical results show PDFD exhibits remarkable performance enhancements 
over many state-of-the-art existing methods. 
\end{abstract}

\section{Introduction}
Semi-supervised learning (SSL) has been widely studied as a leading technique for utilizing 
abundant unlabeled data to reduce the reliance of deep learning models
on extensively labeled datasets \cite{tarvainen2017mean,laine2017temporal}.
Traditional SSL methodologies, operate under a crucial yet often unrealistic assumption: the set of classes encountered during training in the labeled set is exhaustive of all possible categories in the dataset \cite{zhu2005semi}. This assumption is increasingly misaligned with the dynamic and unpredictable nature of real-world data, where new classes can emerge without being labeled, creating a critical gap in the model's knowledge and adaptability \cite{bendale2015towards}. 
This gap underscores the necessity for an Open-World SSL (OW-SSL) setup 
\cite{cao2022open}, 
where the unlabeled data are not only from the classes observed in the labeled data
but also cover novel classes that are previously unseen. 
The investigation of OW-SSL
is essential for maintaining the efficacy and relevance of machine learning models in real-world applications, where encountering new classes is not an exception but a norm.

Diffusion models (DM), initially inspired by thermodynamics \cite{sohl2015deep}, have gained significant popularity, particularly in the realm of generative models \cite{yang2023diffusion,luo2022understanding}. Their application has yielded remarkable success, outperforming established generative models like Variational Autoencoders (VAEs) \cite{kingma2013auto} and Generative Adversarial Networks (GANs) \cite{goodfellow2014generative}, especially in the domain of image synthesis \cite{rombach2022high}. Ongoing developments in DM have led to advancements such as higher-resolution image generation \cite{ho2020denoising}, accelerated training processes \cite{song2021denoising}, and reduced computational costs \cite{rombach2022high}. 
Beyond image generation, 
recent efforts on diffusion models explore their 
application in image classification, incorporating roles as a zero-shot classifier \cite{clark2023texttoimage,li2023diffusion}, integration into SSL frameworks \cite{you2023diffusion,ho2023diffusion}, and enhancing image classification within meta-training phases \cite{du2023protodiff}. This highlights the considerable extensibility of diffusion models.

In this paper, we introduce a novel Prompt-Driven Feature Diffusion (PDFD) approach for 
Open-World Semi-Supervised Learning (OW-SSL), 
specifically designed to overcome the inherent challenges associated with the absence of labeled instances for novel classes in OW-SSL. 
Our approach harnesses the strengths of diffusion models
to enhance effective feature representation learning 
from labeled and unlabeled data 
through instance feature denoising
guided by predicted class-discriminative prompts. 
Recognizing the computational demands of traditional diffusion processes, 
the adopted feature-level diffusion strategy
offers enhanced efficiency and scalability compared to its image-level counterpart. 
Furthermore, feature-level diffusion operates in a representation space where the data is typically more abstract and generalizable, 
allowing the model to utilize the organized information present in labeled data and simultaneously adapt to new classes found within unlabeled data.
A key aspect of PDFD is 
using class prototypes as prompts for
the diffusion process. 
This choice is motivated by the generalizability of prototypes to novel, unseen classes, helping knowledge transfer from seen classes to unseen classes which is crucial in OW-SSL. 
Furthermore, we incorporate a distribution-aware pseudo-label selection strategy during 
semi-supervised training, ensuring proportionate representation across all classes. 
In addition, PDFD uses a class-conditional adversarial 
learning loss \cite{mirza2014conditional} 
to align the prompt-driven features generated by the diffusion process 
with class-conditional real data features, 
reinforcing the guidance of class prototypes for the diffusion process.
This integration effectively bridges SSL classification and adversarial learning, 
leveraging the diffusion model to enhance the fidelity of feature representation in relation to specific classes.
To empirically validate our approach, we conduct extensive experiments across multiple benchmarks in SSL, Open Set SSL, Novel Class Discovery (NCD), and OW-SSL. The results demonstrate that the proposed PDFD model not only outperforms various comparison methods but also achieves state-of-the-art performance in these domains. 
The key contributions of this work can be summarized as follows:
\begin{itemize}
\item We introduce a novel Prompt-Driven Feature Diffusion (PDFD) approach for OW-SSL, 
  which enhances the fidelity and generalizability of feature representation for respective classes
  by leveraging the strengths of diffusion models with properly designed prompts. 
\item 
We deploy a class-conditional adversarial loss to 
support feature-level diffusion model training, 
strengthening the guidance of class prototypes for the diffusion process.
\item 
We utilize a distribution-aware pseudo-label selection strategy, 
ensuring balanced class representation within an SSL framework, 
while class-prototypes are computed on selected instances based on prediction reliability.
\item Our comprehensive empirical results 
demonstrate the superiority of PDFD over a range of SSL, Open-Set SSL, NCD, and OW-SSL methodologies.
\end{itemize}

\section{Related Works}
\subsection{Semi-Supervised Learning} 
\paragraph{Traditional Semi-Supervised Learning (SSL)} 
Traditional SSL 
has focused on training with both labeled and unlabeled data from seen classes,
and classifying
unseen test examples into these ground-truth classes. 
Deep SSL, which applies SSL techniques to deep neural networks, 
can be categorized into entropy minimization methods such as ME \cite{grandvalet2004semi}, 
consistency regularization methods such as Tempral-Ensemble \cite{laine2017temporal} and Mean-Teacher \cite{tarvainen2017mean}, and holistic methods like FixMatch \cite{sohn2020fixmatch}, MixMatch \cite{berthelot2019mixmatch} and ReMixMatch \cite{berthelot2020remixmatch}. 
However, these approaches face challenges when training data includes unlabeled examples from unseen classes. 

\paragraph{Open-Set Semi-Supervised Learning}  
Open-set SSL enhances conventional SSL by recognizing the existence of unseen class examples within the training data while maintaining {\em the premise that unseen classes in the test examples are supposed to just be detected as outliers}. The primary aim in this context is to diminish the detrimental impact that data from unseen classes might have on the classification performance of seen classes. 
To tackle this unique challenge, several recent methodologies have 
employed distinctive strategies
for managing unseen class data.
Specifically, DS3L \cite{guo2020safe} addresses this issue by assigning reduced weights to unlabeled data from unseen classes, while CGDL \cite{sun2020conditional} focuses on improving data augmentation and generation tasks by leveraging conditional constraints to guide the learning and generation process. 
OpenMatch \cite{cao2022open}  employs one-vs-all (OVA) classifiers for determining the likelihood of a sample being an inlier, setting a threshold to identify outliers. However, a common limitation of these approaches is their inability to classify examples from unseen classes.

\paragraph{Novel Class Discovery (NCD)} 
In this setting, 
training data contains labeled examples from seen classes and unlabeled examples from novel unseen classes. 
Distinct from open-set SSL, NCD aims to recognize and classify both seen and unseen classes in the test set.
This problem set-up,  first introduced in \cite{han2019learning}, 
has developed into various methodologies, primarily revolving around a two-step training strategy. Initially, an embedding is learned from the labeled data, followed by a fine-tuning process where clusters are assigned to the unlabeled data \cite{hsu2018learning,han2019learning,fini2021unified}.
A key feature in NCD is the use of the Hungarian algorithm \cite{kuhn1955hungarian} for aligning classes in the labeled data. For instance, Deep Transfer Clustering (DTC) \cite{han2019learning} harnesses deep learning techniques for transferring knowledge between labeled and unlabeled data, aiding in the discovery of novel classes. Another approach, RankStats \cite{han2019automatically}, utilizes statistical analysis of data features to identify new classes.

\paragraph{Open World Semi-Supervised Learning} 
Distinct from NCD, 
OW-SSL encompasses labeled training data from the seen classes and unlabeled training data 
from both the seen and novel unseen classes, 
offering the capacity of exploiting the abundant unlabeled data from seen classes 
that are frequently available in real-world applications. 
As it has just been introduced recently \cite{cao2022open}, 
the potentials of OW-SSL have yet to be fully explored,  
and very few methods have been developed to address its unique challenges. 
ORCA \cite{cao2022open} implements a cross-entropy loss function with an uncertainty-aware adaptive margin,  aiming to reduce the disproportionate impact of the seen (known) classes during the initial phases of training. 
NACH \cite{guo2022robust} brings
instances of the same class in the unlabeled dataset closer
together based on inter-sample similarity. 

\subsection{Diffusion Models}
\paragraph{Diffusion Probabilistic Models (DMs)} 
Originating from principles in thermodynamics, the stochastic diffusion processes were first introduced to data generation in DMs \cite{sohl2015deep}. 
A notable advancement in recent research is Denoising Diffusion Probabilistic Models (DDPMs) proposed in \cite{ho2020denoising}. DDPMs introduce a noise network that learns to predict a series of noise, enhancing the efficiency of DMs in generating high-quality image samples. 
Additionally, Denoising Diffusion Implicit Models (DDIM) were introduced, building upon DDPMs by incorporating a non-Markovian diffusion process, resulting in an acceleration of the generative process \cite{song2021denoising}. Latent Diffusion Models (LDMs) extend the diffusion process to the latent space, enabling DMs to be trained with more efficiency and on limited computational resources \cite{rombach2022high}. They also introduced a cross-attention mechanism to DMs, providing the ability to incorporate conditional information in image generation.
Nevertheless, training diffusion models in generating images is computationally intensive.  
\paragraph{Diffusion Models on Image Classification}
Diffusion Models on Image Classification is a newly emerging area that explores the potential of applying diffusion models to classification tasks. Both \cite{clark2023texttoimage} and \cite{li2023diffusion} consider the diffusion model as a zero-shot classifier. 
\cite{clark2023texttoimage} exploits pre-trained diffusion models and CLIP \cite{radford2021learning}. 
This approach involves generating image samples using text input, scoring, and classifying the image samples. 
Meanwhile, \cite{li2023diffusion} classifies image samples within the noise space. 
Exploring the application of diffusion models in semi-supervised learning tasks,
\cite{ho2023diffusion} learns
image classifier using pseudo-labels generated from the diffusion models. 
\cite{you2023diffusion} uses the diffusion model as a denoising process to obtain bounding box outputs for pseudo-label generation in semi-supervised 3D object detection. 
\cite{du2023protodiff} introduces the concept of prototype-based meta-learning to diffusion models in image classification. 
During the meta-training phase,
it leverages a task-guided diffusion model to gradually generate prototypes,
providing efficient class representations.


\section{Method}
\begin{figure*}[t]
  \centering
 \includegraphics[scale=0.80]{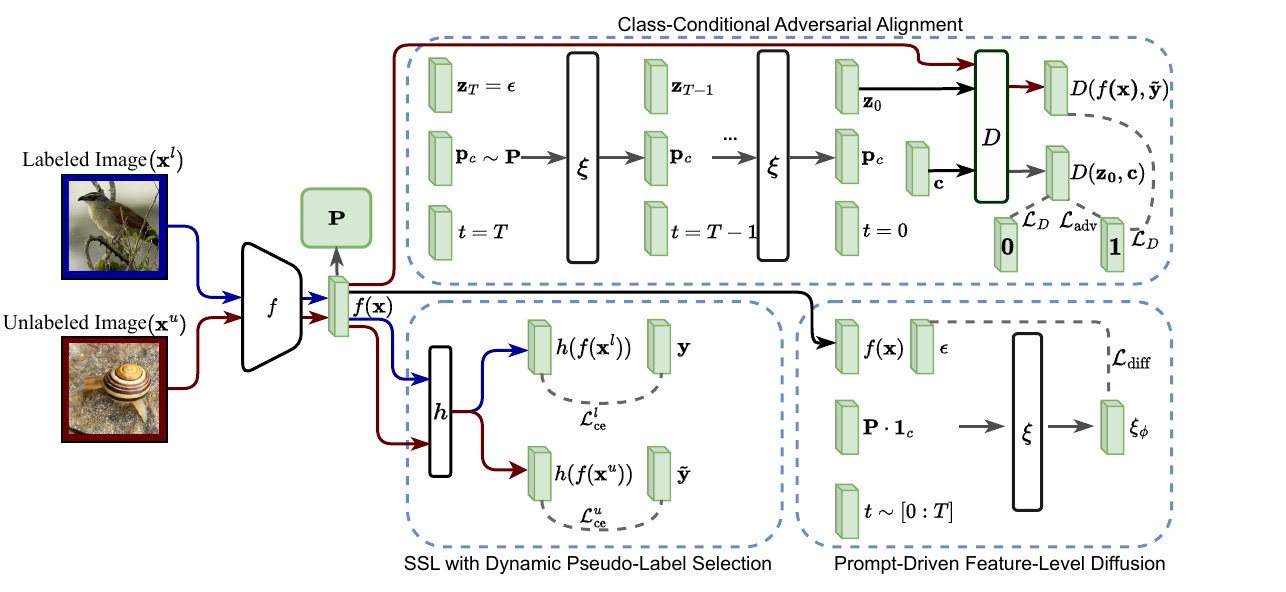} 
	 \caption{The proposed PDFD framework trained on $\mathcal{D}_l$, $\mathcal{D}_u$. 
    The feature encoder $f$ takes as input the labeled data and unlabeled data to generate their learned embeddings. The embeddings of the labeled and unlabeled samples are used to calculate the class prototypes which are used as prompts for the diffusion model. The diffusion model, guided by the loss $\mathcal{L}_{\text{diff}}$, predicts the noise $\xi_\phi$ from noisy features. Concurrently, the classifier $h$ and encoder $f$ are trained, aiming to minimize the supervised loss $\mathcal{L}_{\text{ce}}^{l}$ and the pseudo-labeling loss $\mathcal{L}_{\text{ce}}^{u}$. Additionally, a class conditional-adversarial training component is integrated, wherein the generator $\xi_{\phi}$ aims to produce feature representations that successfully mislead the discriminator $D_{\psi}$, assessed by the adversarial loss $\mathcal{L}_{\text{adv}}$, into categorizing them as real features.
}
   \label{fig:approach}
\vskip -.15in
\end{figure*}
\subsection{Problem Setup}
We consider the following OW-SSL setting. 
The training data comprises
a labeled set $\mathcal{D}_l = \{(\rvx^l_i, \rvy^l_i)\}_{i=1}^{N^l}$ 
with $N^l$ instances, each paired with a corresponding one-hot label vector $\rvy^l_i$ , 
and an unlabeled set 
$\mathcal{D}_u = \{(\rvx^u_i)\}_{i=1}^{N^u}$ with $N^u$ instances. 
The set of classes present in the labeled set are referred to as seen classes, 
denoted as $\mathcal{Y}_s$, 
while the unlabeled data are sampled from a comprehensive set of classes $\mathcal{Y}$,
which includes both the seen classes $\mathcal{Y}_s$ and additional unseen novel classes $\mathcal{Y}_n$,
such that $ \mathcal{Y} = \mathcal{Y}_s \cup \mathcal{Y}_n$. 
The core challenge of OW-SSL is to learn a classifier from the training data 
that can accurately categorize an unlabeled test instance 
to any class in $\mathcal{Y}$.
We aim to learn a deep classification model that 
comprises a feature extractor $f$, parameterized by $\theta_{\text{feat}}$, 
which maps the input data samples from the original input space
$\mathcal{X}$ into a high level feature space $\mathcal{Z}$,
and a linear probabilistic classifier $h$, parameterized by $\theta_{\text{cls}}$. 
The collective parameters of the deep classification model ($h\circ f$) are represented by 
$\theta = \theta_{\text{feat}} \cup \theta_{\text{cls}}$.

\subsection{Diffusion Model Preliminaries}
Diffusion probabilistic models, often simply referred to as ``diffusion models" \cite{sohl2015deep,ho2020denoising}, are a type of generative model characterized by a distinct Markov chain framework.
The diffusion model comprises two primary processes: the forward process and the reverse process. 
The forward process (diffusion process) consists of 
a forward diffusion sequence, 
denoted by $q(\rvx_t | \rvx_{t-1})$, which 
represents a Markov chain that incrementally introduces Gaussian noise at each timestep $t$, starting from an initial clean sample (e.g., image) $\rvx_0 \sim q(\rvx_0)$. 
The forward diffusion process is described mathematically as:
\begin{equation} 
q(\rvx_T |\rvx_0) := \prod_{t=1}^{T} q(\rvx_t | \rvx_{t-1}), 
\end{equation}
where each step is defined via a Gaussian distribution:
\begin{equation} 
q(\rvx_t | \rvx_{t-1}) := \mathcal{N}(\rvx_t; (1 - \beta_t)\rvx_{t-1}, \beta_t\mathbf{I}), 
\end{equation}
with ${\beta}_t$ representing a predefined variance schedule. 
By introducing $\alpha_t := 1 - \beta_t$ and $\bar{\alpha}_t := \prod_{s=1}^{t} \alpha_s$, 
one can succinctly express the diffused sample at any timestep $t$ as:
\begin{equation} 
\rvx_t = \sqrt{\bar{\alpha}_t}\rvx_0 + \sqrt{1 - \bar{\alpha}_t}\epsilon, 
\end{equation}
where $\epsilon$ is a standard Gaussian noise, $\epsilon \sim \mathcal{N}(0, \mathbf{I})$.

Due to the intractability of directly reversing the forward diffusion process, $q(\rvx_{t-1} | \rvx_t)$, the model is trained to approximate this reverse process through parameterized Gaussian transitions, denoted as $p_\phi(\rvx_{t-1} | \rvx_t)$, with $\phi$ as the model parameters. Consequently, the reverse diffusion is modeled as a Markov chain starting from a noise distribution $\rvx_T \sim \mathcal{N}(0, \mathbf{I})$, and is defined as:
\begin{equation} 
p_\phi(\rvx_{0:T}) := p_\phi(\rvx_T) \prod_{t=1}^{T} p_\phi(\rvx_{t-1} | \rvx_t), 
\end{equation}
where the transition probabilities are given by:
\begin{align} 
	p_\phi(\rvx_{t-1} | \rvx_t) &= \mathcal{N}(\rvx_{t-1}; \mu_\phi(\rvx_t, t), \sigma^2_t \mathbf{I}), 
\\
	\mbox{with}\;\,	\mu_\phi(\rvx_t, t) &= \frac{1}{\sqrt{\alpha_t}} \left( \rvx_t - \frac{1 - \alpha_t}{\sqrt{1 - \bar{\alpha}_t}} \xi_{\phi}(\rvx_t, t) \right)
\end{align}
where $\xi$ is the diffusion model parameterized by $\phi$, predicting the added noise. 
In this context, the diffusion model is trained using an objective function defined as follows: 
\begin{equation}   
 \mathcal{L}_\phi = \mathbb{E}_{t, \rvx_0, \epsilon} \left[ \left\| \epsilon - \xi_{\phi}\left(\sqrt{\bar{\alpha}_t} \rvx_0 + \sqrt{1 - \bar{\alpha}_t} \epsilon, t\right)\right\|^2 \right] \end{equation}

\subsection{Proposed Method}

In this section, we outline the proposed Prompt-Driven Feature Diffusion (PDFD) approach 
for OW-SSL. 
We present the method within a semi-supervised learning framework,
where cross-entropy losses on the labeled data and the dynamically selected unlabeled data 
are jointly minimized. 
The key aspect of PDFD is to jointly train 
a feature-level diffusion model with class prototypes as prompts
and the classification model 
through the minimization of a diffusion loss. 
This component is crucial for 
enhancing SSL by leveraging the strengths of diffusion models, 
ensuring semantic distinction and generalizability from the seen to the unseen classes. 
Furthermore, we incorporate a class-conditional adversarial loss 
to align the generated data from the diffusion model 
with the pseudo-labeled real data in the feature space $\mathcal{Z}$,
improving the alignment of feature representation for respective classes. 
The overall framework of PDFD is shown in Figure~\ref{fig:approach}.
Further elaboration will be provided below. 

\subsubsection{SSL with Dynamic Pseudo-Label Selection}

We perform semi-supervised learning over the entire class set $\mathcal{Y}$
by minimizing the cumulative loss over the labeled and unlabeled training data
to learn the parameters $\theta$ of the classification model.
For the labeled data in $\mathcal{D}_l$, we employ the following standard cross-entropy loss: 
\begin{equation}
\label{eq:sup-loss}
	\mathcal{L}_{\text{ce}}^l(\theta)  = \mathbb{E}_{(\rvx^l_i,\rvy^l_i) \in \mathcal{D}_l} [\ell_{ce} (\rvy^l_i, h_{\theta_{\text{cls}}}(f_{\theta_{\text{feat}}}(\rvx^l_i)))]
\end{equation}
where $\ell_{ce}$ denotes the cross-entropy loss function.

For the unlabeled data in $\mathcal{D}_u$, 
we initially produce their pseudo-labels 
using K-means clustering. 
Then in each following training iteration, 
the current classification model is utilized 
to predict the pseudo-labels of each unlabeled instance $\rvx_i^u$ as follows: 
\begin{equation}
\label{eq:pseudo}
	\mathbf{\widehat{y}}_i=h_{\theta_{\text{cls}}}(f_{\theta_{\text{feat}}}(\widehat{\rvx}_i^u))
\end{equation}
where $\mathbf{\widehat{y}}_i$ denotes a soft pseudo-label vector---i.e., 
the predicted class probability vector with length $|\mathcal{Y}|$;
$\widehat{\rvx}_i^u$ denotes a weakly augmented version of instance $\rvx_i^u$.  
By using weak augmentation, we aim to capture the underlying structure of 
the unlabeled data while minimizing the impact of potential noise or distortions.
The corresponding one-hot pseudo-label vector $\widetilde{\mathbf{y}}_i$ 
can be produced from $\mathbf{\widehat{y}}_i$ by setting 
the entry with the largest probability as 1 while keeping other entries as 0s. 

Moreover, in order to minimize the impact of noisy pseudo-labels
and ensure a proportionate representation of all classes in the unlabeled data, 
we propose to dynamically select confident pseudo-labels 
to produce a distribution-aware subset of pseudo-labeled instances for model training. 
Specifically, for each class $c\in\mathcal{Y}$,
we choose a subset of instances, $\mathcal{C}_c$, 
with confidently predicted pseudo-labels via a threshold $\tau$:
\begin{equation}
\!\!\!\!
\label{eq:confident-pl}
\mathcal{C}_c = \{ \rvx_i^u\!\in\!\mathcal{D}_u |\mathbbm{1}\big( \max(\mathbf{\widehat{y}}_i)\! >\! \tau 
	\, \land  \, 
	 \argmax\nolimits_j \mathbf{\widehat{y}}_{ij}\! =\! c\big) \} 
\end{equation}
where the indicator function $\mathbbm{1}\big(\cdot)$ presents the condition for instance selection. 
The minimum number of instances selected for each class 
can then be determined as $N_m = \min_c |\mathcal{C}_c|$.

To ensure a well-proportioned consideration for all the classes $\mathcal{Y}$,
we finally choose the top $N_m$ instances from each pre-selected subset $\mathcal{C}_c$
based on the predicted pseudo-label scores, $\max(\mathbf{\widehat{y}}_i)$,
and form a selected pseudo-labeled set 
$\mathcal{Q}=\{ (\rvx_i,\mathbf{\widetilde{y}}_i),\cdots\}$ with size $N_m\times |\mathcal{Y}|$.

The training loss on the unlabeled data is then computed as the cross-entropy loss
on the confidently pseudo-labeled instances in $\mathcal{Q}$:
\begin{equation}
\label{eq:pl-loss}
	\mathcal{L}_{\text{ce}}^u(\theta) = \mathbb{E}_{(\rvx_i, \mathbf{\Tilde{y}}_i) \in \mathcal{Q}} 
	[\ell_{\text{ce}}(\mathbf{\Tilde{y}}_i,h_{\theta_{\text{cls}}}(f_{\theta_{\text{feat}}}(\rvx_i^u)))]
\end{equation}
%

\subsubsection{Class-Prototype Computation}

Prior to introducing the key feature-level diffusion component, 
we first compute the class-prototypes that will be adopted as essential prompts for 
guiding the diffusion process. 

In particular, class prototypes are derived from the feature embeddings produced 
from the deep feature extractor $f$ based on the (predicted) class labels.
They hence encapsulate the core characteristics of classes
in the high level semantic feature space $\mathcal{Z}$
that are generalizable to novel categories.

For the seen classes in $\mathcal{Y}_s$, 
we calculate the class prototypes as average feature representations 
of the labeled data for each class, 
providing a stable reference point for the whole class set $\mathcal{Y}$.
Specifically for each class $s \in \mathcal{Y}_s$, 
we compute its class prototype vector $\mathbf{p}_{s}$ as follows: 
\begin{equation}
\label{eq:labeled-p}
    \mathbf{p}_{s}  = \mathbb{E}_{(\rvx^l_i,\mathbf{y}_i^l)\in \mathcal{D}_l} 
	\left[ \mathbbm{1}\big( \argmax\nolimits_j\mathbf{y}_{ij}^l = s \big) f_{\theta_{\text{feat}}}(\rvx^l_i)\right]
\end{equation}
where the indicator function $\mathbbm{1}(\cdot)$ selects the instances 
that satisfy the given conditions---belonging to class $s$ in this case. 

For the unseen novel classes in $\mathcal{Y}_n$, 
the prototypes are computed differently to account for the uncertainty during the discovery of new classes
on unlabeled data.
Specifically, for each class $n\in\mathcal{Y}_n$, 
its class prototype vector  $\mathbf{p}_{n}$ is computed 
as the average feature representation of the unlabeled instances
whose pseudo-labels are confidently predicted as class $n$:
\begin{equation}
 \label{eq:unlabeled-p}
     \mathbf{p}_{n}  = \mathbb{E}_{\rvx^u\in \mathcal{D}_u} 
\left[ \mathbbm{1}\big( \max(\mathbf{\widehat{y}}_i) > \tau \, \land  \, 
	 \argmax\nolimits_j \mathbf{\widehat{y}}_{ij} = n \big) f_{\theta_{\text{feat}}}(\rvx^u_i)\right]
\end{equation}
where the threshold $\tau$ is used to filter out non-confident predictions
and reliably identify novel unseen classes in the unlabeled data. 
By putting all these class prototypes together, 
we can form a class prototype matrix 
$\mathbf{P}=[\mathbf{p}_1,\cdots,\mathbf{p}_{\footnotesize|\mathcal{Y}|}]$,
whose each column contains a class prototype vector.

\subsubsection{Prompt-Driven Feature-Level Diffusion}
Traditional diffusion processes, while powerful, are often computationally intensive and time-consuming, particularly when applied directly to high-dimensional data such as images. By transposing the diffusion process to the feature level, we significantly reduce the computational burden, enabling faster training of the diffusion model and scalability of PDFD to large datasets. 
In addition, feature-level diffusion focuses on 
the high-level representation space where the data is often more abstract and generalizable,
while the semantic aspects of the data captured in this space are more relevant and informative for classification. 
Image-level diffusion might inadvertently emphasize pixel-level details that are less important for understanding the underlying class or concept. 
By operating at the feature level, the model can leverage the global and structural information 
to distinguish novel classes from seen classes. 

To leverage the strengths of the diffusion model for class distinction and recognition,
we introduce the class prototypes as an additional input to the standard diffusion model 
$\xi_\phi$, functioning as class-distinctive prompts for feature diffusion. 
Specifically, 
the model is tasked with predicting the added noise $\epsilon$ 
based on a noisy input feature vector, the class-specific prompt, and the current time step $t$:
\begin{equation}
	\xi_\phi=\xi_\phi(\sqrt{\bar{\alpha}_t} f_{\theta_{\text{feat}}}(\rvx_i) + \sqrt{1 - \bar{\alpha}_t} \epsilon, \mathbf{P}  \cdot \mathbf{1}_{c_i}  , t)
\end{equation}
where $\mathbf{1}_{c_i}$ denotes a one-hot vector that indicating the predicted class
of the corresponding input $\rvx_i$,
such that 
$c_i = \argmax_j\; h_{\theta_{\text{cls}}}(f_{\theta_{\text{feat}}}(\rvx_i))[j]$;
while $\mathbf{P}  \cdot \mathbf{1}_{c_i}$ chooses the corresponding class prototype vector
as the prompt input. 
Same as in the standard diffusion model, 
the term $ \bar{\alpha}_t $ is a pre-defined variance schedule, 
and $ \epsilon$ is a noise variable sampled from the normal distribution.
Following \cite{du2023protodiff}, we employ a transformer-based diffusion model for $\xi_{\phi}$.   

We jointly train the diffusion model $\phi$ and the 
classification model $\theta$ (feature extractor $\theta_{\text{feat}}$ and classifier $\theta_{\text{cls}})$
over all the labeled and unlabeled training instances by minimizing the following diffusion loss:
\begin{equation}
\label{eq:loss-diff}
	\mathcal{L}_{\text{diff}}(\phi, \theta) 
	= \mathbb{E}_{\rvx_i \in \mathcal{D}_l\cup\mathcal{D}_u}\mathbb{E}_{t \sim [0:T]}
	[\lVert \epsilon - \xi_\phi \rVert^2 ]
 \end{equation}
The loss essentially measures the discrepancy between the added noise $\epsilon$ and the prediction of 
the generative diffusion model, 
guiding both the feature extractor and the diffusion model to produce feature representations 
that are coherent with the class prototypes and therefore suitable for both seen and unseen class identification.
%

\subsubsection{Class-Conditional Adversarial Alignment}
The data generation in our PDFD model 
is depicted through a reverse diffusion process, where we transform a random 
noise vector $\epsilon$ in a sequence of $T$ steps into meaningful feature vectors in the high-level 
feature representation space $\mathcal{Z}$,
guided by a class prototype based prompt. 
The process is mathematically represented as:
\begin{equation}
 \label{eq:revers-diffusion}
\mathbf{z}_{t-1}\!=\!
\begin{cases}
\epsilon & \text{if } t = T, \\
\frac{1}{\sqrt{\alpha_t}} \left( \mathbf{z}_t \!-\! \frac{1 - \alpha_t}{\sqrt{1 - \bar{\alpha}t}} \cdot \xi_{\phi}(\mathbf{z}_t, \mathbf{p}_c, t) \right) & \text{if } t <T
\end{cases}
\end{equation}
where $\mathbf{z}_t$ denotes the 
diffused feature embedding vector at time step $t$. 
For simplicity, we define this reverse diffusion process 
as a generative function $g_\phi(\epsilon, T, \mathbf{p}_c)$, 
which takes the initial noise vector $\epsilon$, the total number of time steps $T$, 
and the prompt $\mathbf{p}_c$ as inputs,
and generates a diffused clean feature vector $\mathbf{z}_0$: 
\begin{equation}
\label{eq:clean-feat}
 \mathbf{z}_0 = g_\phi(\epsilon, T, \mathbf{p}_c)   
\end{equation}
Here, $g$ conveniently encapsulates the iterative reverse diffusion process, 
transforming the initial noise $\epsilon$ into the refined feature representation $\mathbf{z}_0$ 
through a sequence of $T$ steps of transformations governed by the specified prompt 
and the dynamics of the diffusion process in Eq.(\ref{eq:revers-diffusion}).

In advancing our model's robustness
and diffusion capacity, 
we propose to align the generated feature vectors 
with the unlabeled real training data in the high-level feature space $\mathcal{Z}$ 
through a class-conditional adversarial loss defined as follows:
\begin{align}
	\!\!\!\!\!\!\!\!\!\!
	&\mathcal{L}_{\text{adv}}(\phi,\psi) = 
	\mathbb{E}_{\mathbf{x} \sim \mathcal{D}_u}[\log D_\psi({f_{\theta_{\text{feat}}}(\mathbf{x})},\mathbf{\Tilde{y}})]
	\nonumber\\ 
	&\quad + \mathbb{E}_{\epsilon \sim \mathcal{N} (0, \mathbf{I}), {c} \sim \mathcal{Y} }
	[\log (1 - D_\psi(g_\phi(\epsilon, T, \mathbf{p}_c), \mathbf{1}_{c}))],
\label{eq:loss-adv}
\end{align}
where $D_\psi$ is a class-conditional discriminator parameterized by $\psi$, 
which tries to maximumly distinguish the feature vectors of the real data
from the generated feature vectors using the reverse diffusion process
given the conditional one-hot label vector. 
This adversarial loss is tailored to refine the model's ability to 
generate class-specific features. 
By playing a minimax adversarial game
between the diffusion model $\phi$ and the discriminator $\psi$, 
\begin{equation}
\label{eq:minimax}
   \min_\phi \max_\psi\; \mathcal{L}_{\text{adv}}(\phi,\psi),  
\end{equation}
this class-conditional adversarial alignment loss 
encourages the diffusion model to generate features that are indistinguishable from real data features, 
enhancing the fidelity of feature representation w.r.t respective classes
across both the seen and unseen classes in $\mathcal{Y}$. 

\subsubsection{Joint Training of PDFD}
Incorporating the SSL losses on both labeled and unlabeled data sets, 
alongside the diffusion and adversarial losses, 
we formulate the joint training objective for our PDFD model as follows:
\begin{equation}
 \min_{\theta,\phi}\max_\psi\; \mathcal{L}_{\text{tr}} = \mathcal{L}^l_{\text{ce}} + \gamma_u  \mathcal{L}_{\text{ce}}^u+ \gamma_{\text{diff}}  \mathcal{L}_{\text{diff}}+ \gamma_{\text{adv}} \mathcal{L}_{\text{adv}}
 \end{equation}
where $\gamma_{u}$, $\gamma_{\text{diff}}$ and $\gamma_{\text{adv}}$ are trade-off hyper-parameters. 


\section{Experiments}
\begin{table*}[h]
\centering
\caption{Classification accuracy (\%) on CIFAR-10, CIFAR-100, and ImageNet-100.}
\setlength{\tabcolsep}{4pt}	
\begin{tabular}{c|c|c|c|c|c|c|c|c|c}
\hline
Classes & Dataset & SSL & \multicolumn{2}{c|}{Open-Set SSL} & \multicolumn{2}{c|}{NCD} & \multicolumn{3}{c}{Open-World SSL}\\
\hline
 & & Fixmatch & DS3L & CGDL  & DTC & RankStats & ORCA& NACH &PDFD (ours)\\
\hline
Seen & CIFAR-10 & 71.5 & 77.6 & 72.3 & 53.9 & 86.6 & 88.2 & 89.5 &$\mathbf{90.2}$\\
 & CIFAR-100 & 39.6 & 55.1 & 49.3 & 31.3 & 36.4 & 66.9 & 68.7& $\mathbf{70.2}$\\
 & ImageNet-100 & 65.8 & 71.2 & 67.3 & 25.6 & 47.3 & 89.1 & 91.0 &$\mathbf{91.3}$\\
 & Average & 59.0 & 68.0 & 63.0 & 36.9 & 56.8 & 81.4 & 83.1&$\mathbf{83.9}$ \\
\hline
Unseen & CIFAR-10 & 50.4 & 45.3 & 44.6 & 39.5 & 81.0 & 90.4 & 92.2 &$\mathbf{93.1}$\\
 & CIFAR-100 & 23.5 & 23.7 & 22.5 & 22.9 & 28.4 & 43.0 & 47.0  &$\mathbf{49.5}$\\
 & ImageNet-100 & 36.7 & 32.5 & 33.8 & 20.8 & 28.7 & 72.1 &75.5& $\mathbf{76.1}$\\
 & Average & 36.9 & 33.9 & 33.6 & 27.7 & 46.0 & 68.5 & 71.6&$\mathbf{72.9}$\\
\hline
All & CIFAR-10 & 49.5 & 40.2 & 39.7 & 38.3 & 82.9 & 89.7 & 91.3&$\mathbf{92.1}$ \\
 & CIFAR-100 & 20.3 & 24.0 & 23.5 & 18.3 & 23.1 & 48.1 &52.1&  $\mathbf{52.9}$\\
 & ImageNet-100 & 34.9 & 30.8 & 31.9 & 21.3 & 40.3 & 77.8 & 79.6& $\mathbf{80.6}$ \\
 & Average & 34.9 & 31.7 & 31.7 & 26.0 & 48.8 & 71.9 & 74.3& $\mathbf{75.2}$\\
\hline
\end{tabular}
\label{tab:classification_accuracy}
\vskip -0.05 in
\end{table*}

\subsection{Experimental Setup}
\paragraph{Datasets} We evaluate our model using established benchmarks in image classification: CIFAR-10, CIFAR-100 \cite{krizhevsky2009learning}, and a subset of ImageNet \cite{deng2009imagenet}. The chosen ImageNet subset encompasses 100 classes, given its expansive size. Each dataset is partitioned such that the first 50\% of the classes are considered 'seen' and the rest as 'novel'. For these seen classes, we label 50\% of the samples and the remainder constitutes the unlabeled set.
 The results presented in this study were obtained from evaluations on an unseen test set, which comprises both previously seen and novel classes, ensuring a comprehensive assessment of the model's performance. We repeated all experiments for 3 runs and reported the average results.
 
\paragraph{Experimental Setup} Following the compared methods \cite{cao2022open,guo2022robust}, we pretrain our model using simSLR \cite{chen2020simple} method. In our experiments with the CIFAR datasets, we chose ResNet-18 as our primary backbone architecture. The training process involves Stochastic Gradient Descent (SGD) with a momentum value set at 0.9 and a weight decay factor of 5e-4. The training duration is 200 epochs, using a batch size of 512. Only the parameters in the final block of ResNet are updated during the training to prevent overfitting.
For the ImageNet dataset, the backbone model selected is ResNet-50 employing standard SGD for training, with a momentum of 0.9 and a weight decay of 1e-4. We train the model for 90 epochs and maintain the same batch size of 512. Across all our experiments, we apply the cosine annealing schedule to adjust the learning rate.
Specifically for PDFD we set $\gamma_u$ to 0.5, $\gamma_{\text{diff}}$ to 1, $\gamma_{\text{adv}}$ to 1, $\tau$ to 0.5 and $T$ to 50.  
Regarding the architecture of the diffusion model, we adopt a transformer-based model in line with the methodology outlined in \cite{du2023protodiff}. 
The discriminator 
consists of three linear layers, with the first two followed by batch normalization and a ReLU activation function. 

\subsection{Comparison Results}
We conducted a comprehensive comparison of our PDFD method with various state-of-the-art SSL methods across different settings, including Fixmatch \cite{sohn2020fixmatch} for standard SSL, DS3L \cite{guo2020safe} and CGDL \cite{sun2020conditional} for open-set SSL, DTC \cite{han2019learning} and RankStats \cite{han2019automatically} for NCD, and ORCA \cite{cao2022open} and NACH \cite{guo2022robust} for OW-SSL. The evaluation included datasets of varying scales, namely CIFAR-10, CIFAR-100 \cite{krizhevsky2009learning}, using Resnet-18 backbone and ImageNet-100 \cite{russakovsky2015imagenet} using Resnet-50 backbone.  The results presented in this study were obtained from evaluations on an unseen test set, which comprises both previously seen and novel classes, ensuring a comprehensive assessment of the model's performance.

The comparative results are presented in Table \ref{tab:classification_accuracy}. 
The results for all classes illustrate method performance in an OW-SSL setting where both seen and unseen classes are included in the test set. Our PDFD method outperforms all comparison methods across all datasets. Notably, on the ImageNet-100 dataset, PDFD exhibits a significant improvement of $1.0\%$ on all classes compared to the previous state-of-the-art method NACH. It also demonstrates a $0.8\%$ margin of improvement over the second-best algorithm on the CIFAR-10 and CIFAR-100 datasets. The results show an overall performance increase of $0.9\%$ on the average of all three datasets. We also evaluated the effectiveness of the methods in classifying unseen classes. On unseen classes, PDFD outperforms the previous best method on all three datasets.
PDFD performs exceptionally well on the CIFAR-100 dataset, surpassing the second-best method with a significant improvement of $2.5\%$ on unseen classes. On both CIFAR-10 and ImageNet-100 datasets, PDFD also surpasses the previous best methods, exhibiting a $0.9\%$ and $0.6\%$ increase in overall performance across all three datasets on unseen classes. Despite the special treatment of novel classes in the unlabeled dataset, 
PDFD also demonstrates strong performance in standard SSL tasks. PDFD outperforms all previous SSL methods, even on standard SSL tasks on seen classes. PDFD exhibits a similar pattern on the CIFAR-100 dataset as on unseen classes, with a significant improvement of $1.5\%$ over the second-best algorithm. PDFD also shows a $0.8\%$ increase compared to the second-best method in the average classification accuracy across all three datasets, demonstrating the best overall performance.

\subsection{Ablation Study}
\subsubsection{Ablation on different prompts}
We conducted an ablation study to investigate employing different types of prompts  
in PDFD. We compared the classification accuracy on the CIFAR-100 dataset with the full PDFD model, which used prototype corresponding to class prediction ($\mathbf{P}.\mathbf{1}_{c}$) as prompts, 
and two ablation variants. (1) ``$h_{\theta_{\text{cls}}}(f_{\theta_{\text{feat}}}(\rvx_i))$", which uses raw probability prediction output for the sample and (2) ``$\mathbf{1}_{c}$"  which uses one-hot encoding of the prediction output from the feature extractor.
The results of the ablation study are presented in Table \ref{tab:prompt-ablation-transposed}. Notably, utilizing prototypes as prompts achieved the highest accuracy among all three variants. Particularly in unseen classes, the use of prototypes significantly improved the classification performance. This finding 
suggests
that class prototypes are a suitable way to implement prompts in our method, especially in enhancing the performance of PDFD in classifying unseen examples.
\begin{table}[t]
\centering
\caption{Ablation Study on the effect of different types of prompt. classification accuracy (\%) on CIFAR-100.}
\setlength{\tabcolsep}{4pt} 

\begin{tabular}{l|c|c|c}
\hline
Prompt & Seen & Unseen & All \\
\hline
$h_{\theta_{\text{cls}}}(f_{\theta_{\text{feat}}}(\rvx_i))$ & 67.2 & 46.1 & 50.8 \\
$\mathbf{1}_{c}$ & 69.2 & 47.8 & 52.0 \\

$\mathbf{P}.\mathbf{1}_{c}$ (PDFD)& $\mathbf{70.2}$ & $\mathbf{49.5}$ & $\mathbf{52.9}$ \\
\hline
\end{tabular}
\label{tab:prompt-ablation-transposed}
\vskip -0.1 in
\end{table}

\begin{table}[t]
\centering
\caption{Ablation Study classification accuracy (\%) on CIFAR-100.}
\setlength{\tabcolsep}{4pt}	
\begin{tabular}{l|c|c|c}
\hline
  &  Seen&Unseen&All \\
\hline
PDFD& $\mathbf{70.2}$ & $\mathbf{49.5}$ & $\mathbf{52.9}$ \\
\hline
$\;- \text{w/o }\mathcal{L}^l_{\text{ce}}$   &57.6& 24.9 & 45.5  \\
$\;- \text{w/o }\mathcal{L}^u_{\text{ce}}$  & 67.9 & 45.6 & 49.3 \\

$\;- \text{w/o }\mathcal{L}_{\text{diff}}$ &67.1 & 46.4 &48.7 \\
$\;- \text{w/o }  \mathcal{L}_{\text{adv}}$ & 68.0 & 46.9 &50.1 \\
$\;- \text{w/o } \mathcal{L}_{\text{adv}} \text{ and } \mathcal{L}_{\text{diff}} $  & 66.6 & 45.2 &47.7 \\
$\;- \text{w/o }\text{Class condition}$ & 68.1 & 47.1 &50.7 \\

\hline
\end{tabular}
\label{tab:prompt-ablation}
\vskip -0.1 in
\end{table}

\begin{figure}[t]

\centering
\begin{subfigure}{0.23\textwidth}
\centering
\includegraphics[width = \textwidth, height=1.2in]{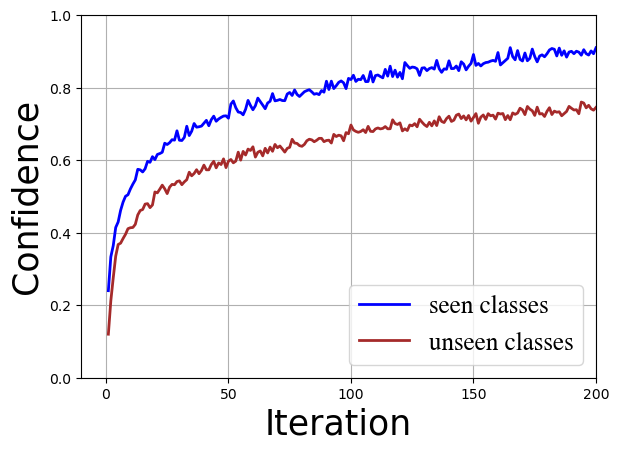}
\caption{Confidence difference between
seen and unseen classes }
\end{subfigure}
\begin{subfigure}{0.23\textwidth}
\centering
\includegraphics[width = \textwidth, height=1.2in]
{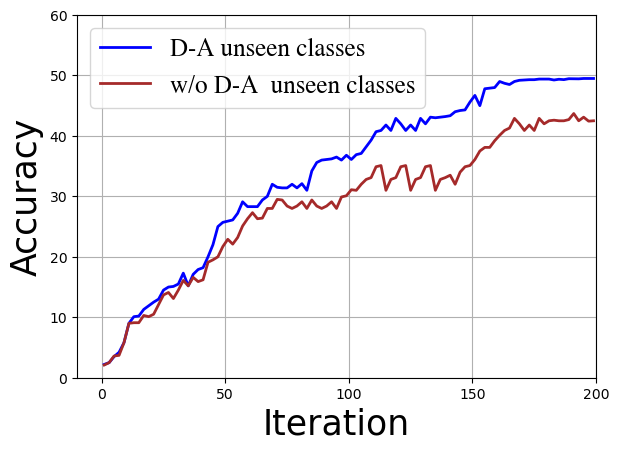}

\caption{ Accuracy of pseudo-labels for
unseen classes. }
\vskip -0.15 in
\end{subfigure}
\caption{Pseudo-Label Selection Analysis. (a) Confidence difference between
seen and unseen classes during the training on CIFAR-100 (b) Effect of distribution-aware pseudo-label selection on learning unseen classes during the training on CIFAR-100. }
\label{fig:pseudo-selection}
\vskip -0.15 in
\end{figure}

\subsubsection{Ablation on different components}
We conducted an ablation study to 
investigate the impact
of different components in PDFD 
on the overall performance. The study focused on classification accuracy using the CIFAR-100 dataset, with six ablation variants:
(1) ``$- \text{w/o }\mathcal{L}^l_{\text{ce}}$'' excluding cross entropy loss on labeled data;
(2) ``$- \text{w/o }\mathcal{L}^u_{\text{ce}}$'' excluding cross entropy loss on unlabeled data;
(3) ``$- \text{w/o }$  $\mathcal{L}_{\text{diff}}$'' excluding diffusion loss, disabling the feature-level diffusion model;
(4) ``$- \text{w/o }  \mathcal{L}_{\text{adv}}$'' excluding adversarial loss, disabling adversarial training;
(5) ``$- \text{w/o } \mathcal{L}_{\text{adv}} \text{ and } \mathcal{L}_{\text{diff}} $'' excluding both adversarial training and diffusion model;
(6) ``$- \text{w/o }  $ Class condition'' excludes the prompt in the diffusion model and class condition in adversarial training.

The ablation study results are presented in Table \ref{tab:prompt-ablation}. PDFD achieves the highest classification accuracy across seen, unseen, and all classes, emphasizing the effectiveness of all model components.
Notably, excluding supervised learning (``$- \text{w/o }\mathcal{L}^l_{\text{ce}}$'') results in the most decreased accuracy. Excluding the diffusion model (``$- \text{w/o }$  $\mathcal{L}_{\text{diff}}$'') significantly lowers accuracy on seen and all classes, emphasizing the importance of this model component. While further excluding adversarial training (``$- \text{w/o } \mathcal{L}_{\text{adv}} \text{ and } \mathcal{L}_{\text{diff}} $'') does not markedly impact seen class accuracy, it does lead to reduced performance on unseen and all classes, supporting the goal of adversarial training to learn indistinguishable pseudo-labels for novel classes.
The exclusion of cross entropy loss on unlabeled data (``$- \text{w/o }\mathcal{L}^u_{\text{ce}}$'') results in a dramatic decrease in model performance on unseen and all classes. This finding supports the significance of each component in contributing to the effectiveness of PDFD.

\subsubsection{Pseudo-Label Selection Analysis}

Figure \ref{fig:pseudo-selection} illustrates the learning analysis of pseudo-labels throughout the training process. As depicted in subfigure (a), it is evident that the seen classes satisfy the confidence condition earlier than the unseen classes. Consequently, this leads to the under-representation of unseen classes in the initial stages of training, culminating in a suboptimal initialization of the model. This early skew towards seen classes can potentially bias the model's learning, impacting its ability to effectively recognize and adapt to the characteristics of the unseen classes as training progresses. In subfigure (b), the positive impact of our proposed component, distribution-aware pseudo-label selection, on the learning of unseen classes is visible. This method effectively addresses the initial imbalance observed in the learning process, enhancing the model's ability to recognize and accurately classify unseen classes. By considering the distribution characteristics of the data, our solution ensures a more equitable representation of classes in the training process, leading to improved model performance and generalization.


\section{Conclusion}
In this paper, we proposed a novel Prompt-Driven Feature Diffusion (PDFD) approach 
to address the challenging setup of Open-World Semi-supervised Learning.
The proposed PDFD approach 
deploys an efficient feature-level diffusion model with class-prototypes as prompts,
enhancing the fidelity and generalizability of feature representation 
across both the seen and unseen classes.
In addition, a class-conditional adversarial loss is further incorporated
to support diffusion model training, strengthening the guidance of class prototypes for the diffusion process.
Furthermore, we also utilized 
a distribution-aware pseudo-label selection strategy 
to ensure balanced class representation for SSL
and reliable class-prototypes computation for the novel classes.
We conducted extensive experiments on several benchmark datasets.
Notably, our approach has demonstrated superior performance
over a set of state-of-the-art methods for SSL, open-set SSL, NCD and OW-SSL.

\clearpage
\bibliographystyle{named}
\bibliography{ijcai24}

\end{document}